\documentclass[letterpaper, 10 pt, conference]{ieeeconf}  

\IEEEoverridecommandlockouts                              

\usepackage{multirow}
\usepackage{graphics} 
\usepackage{epsfig} 
\usepackage{mathptmx} 
\usepackage{times} 
\usepackage{amsmath} 
\usepackage{amssymb}  
\usepackage[utf8]{inputenc}
\usepackage{subcaption}
\usepackage{pdfpages}
\usepackage{paralist}
\usepackage[font=small,labelfont=bf]{caption}
\usepackage{array}
\usepackage{booktabs}
\usepackage{lipsum}
\usepackage{caption}
\usepackage{siunitx}
\usepackage{filecontents}
\usepackage{hyperref}
\usepackage{float}
\usepackage{cite}
\usepackage{makecell}

\usepackage{tabularx}

\title{\LARGE \bf
LBAP: Improved Uncertainty Alignment of LLM \\ Planners using Bayesian Inference
}

\author{James F. Mullen Jr$^1$, and Dinesh Manocha$^1$ \\
{\small{Supplemental version including Code, Video, and Appendix is available at \url{https://gamma.umd.edu/LBAP/}}}
\thanks{
$^1$The authors are associated with the University of Maryland, College Park, USA
{\tt\small mullenj@umd.edu, dmanocha@umd.edu}}%
}
\newcolumntype{M}[1]{>{\centering\arraybackslash}m{#1}}

\begin{document}
\bstctlcite{IEEEexample:BSTcontrol}
\maketitle
\thispagestyle{empty}
\pagestyle{empty}

\begin{abstract} 
Large language models (LLMs) showcase many desirable traits for intelligent and helpful robots. However, they are also known to hallucinate predictions. This issue is exacerbated in robotics where LLM hallucinations may result in robots confidently executing plans that are contrary to user goals or relying more frequently on human assistance. In this work, we present LBAP, a novel approach for utilizing off-the-shelf L\underline{L}Ms, alongside \underline{B}ayesian inference for uncertainty \underline{A}lignment in robotic \underline{P}lanners that \textit{minimizes hallucinations and human intervention}. Our key finding is that we can use Bayesian inference to more accurately calibrate a robots confidence measure through accounting for both scene grounding and world knowledge. This process allows us to mitigate hallucinations and better align the LLM's confidence measure with the probability of success. Through experiments in both simulation and the real world on tasks with a variety of ambiguities, we show that LBAP significantly increases success rate and decreases the amount of human intervention required relative to prior art. For example, in our real-world testing paradigm, LBAP decreases the human help rate of previous methods by over 33\% at a success rate of 70\%.
\end{abstract}

\section{Introduction}
 \begin{figure}[t]
     \centering
     \includegraphics[width= 0.85\linewidth]{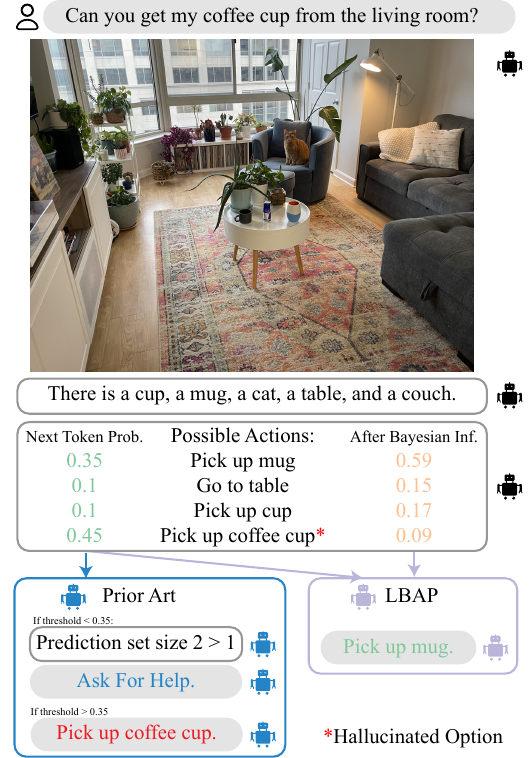}
     \caption{LLMs are not grounded in the real world and will frequently hallucinate. Additionally, when provided with an ambiguous instruction, the LLM must know when to ask users for help. We present LBAP, a novel approach for calibrating LLM confidence with Bayesian inference to better detect hallucinations and resolve ambiguities, before asking a user for help when necessary. 
     Note, in the provided example, `pick up coffee cup' is a hallucination caused by the users phrasing of their command. No such specific object was grounded in the perception information. In LBAP, Bayesian inference incorporates this fact into the final probabilities, allowing us to mitigate this case. }
     \label{fig:coverimg}
 \end{figure}
 
 Imagine you have a home robot and you want it to bring you your coffee cup. As you tell the robot your instruction, it should comprehend your goal, no matter how you phrase the instruction, and ideally complete the task without further clarification. Now imagine there are multiple coffee cups on the counter, the robot may need to ask you for help determining which one is yours. Finally, imagine that there is one coffee cup, but many other objects on the counter that may distract the robot. You would expect the robot to get the coffee cup, without asking for your assistance. If the robot asked you for help, brought you the wrong item, or failed due to some hallucination, you would be inclined to question its ability and potentially be less likely to trust it in the future. There will always be uncertainty in the unstructured and novel environments these types of robots operate in, but they must operate reliably and intelligently nonetheless.

 Recent approaches that leverage large language models (LLMs) for planning \cite{ichtersaycan2023, dorbalaCanEmbodiedAgent2023, renRobotsThatAsk2023}
 have demonstrated an ability to navigate these types of complex environments with a higher success rate than prior methods, while also responding properly to natural and unstructured language instructions. Furthermore, each new generation of the language model, such as GPT-3 to GPT-4, greatly improves model capabilities and intelligence, and subsequently performance on these robotics tasks. However, a significant challenge with all LLMs, new or old, is their tendency to present an incorrect answer confidently, or \textit{hallucinate} \cite{guanHallusionBenchAdvancedDiagnostic2023, dhuliawalaChainofVerificationReducesHallucination2023}. 
 

Additional challenges arise when the LLM is embodied and must interact with users directly.
Significant uncertainty is present as human-provided instructions can be ambiguous, in some cases causing further hallucinations. Completing the wrong action or hallucinating and failing could hurt user trust or even be dangerous. One method of mitigating these issues is to ask users for help or clarifications when needed \cite{renRobotsThatAsk2023, huangInnerMonologueEmbodied2022, dialfred, cvdn}. However, most prior work does not ask for clarification from users in uncertain or ambiguous situations, or does so via extensive manual programming, oftentimes excessively relying on seeking assistance \cite{huangInnerMonologueEmbodied2022}. More recently, KnowNo \cite{renRobotsThatAsk2023} frames the task of when a robot should ask for help as \textit{uncertainty alignment}, and outlines the main challenges for robots that ask for help: (i) \textit{calibrated confidence} -- the robot should seek sufficient help to ensure a probability of task success, and (ii) \textit{minimal help} -- the robot should minimize the amount of help it seeks. However, KnowNo primarily focuses on creating a means of evaluation for the \textit{uncertainty alignment} task and provides a basic baseline. In contrast, we aim to improve the confidence measure, increase the success rate and minimize human intervention, moving towards more helpful and less annoying robot partners.

\textbf{Main Contributions}:
 We introduce LBAP, a novel approach for calibrating LLM confidence with Baysian inference. Our key finding is that using the LLM next token log probability, as proposed by KnowNo \cite{renRobotsThatAsk2023}, is far from enough to adequately calibrate confidence in a real-world setting; both grounding potential actions and applying world knowledge are necessary. Bayesian inference allows us to do just this by conditioning the probability of each possible action on both the grounding in the scene and the available world knowledge.
 Intuitively, we can think of our Bayesian inference step as integrating the potential for an action to be both possible and safe in the given scene into the confidence measure for each potential action. These steps minimize hallucinations, improve safety, and more accurately calibrate uncertainty. By handling these issues, LBAP can be used to increase the task success rate and minimize the frequency of human intervention. An additional advantage of LBAP is that it does not require extensive training (as suggested in SayCan \cite{ichtersaycan2023}), which is difficult to do without overfitting for extremely diverse and data-limited tasks like those found in home robotics. 

The main contributions of our work include:
\begin{enumerate}
    \item We introduce LBAP, a novel approach for uncertainty alignment that uses Large \underline{L}anguage Models and \underline{B}ayesian inference to better \underline{A}lign model confidence with task success for high-level robotic \underline{p}lanners.
    \item We specifically propose using Bayesian inference to combine scene grounding and world knowledge with a prior coming from the LLM next token log probabilities.
    \item We evaluate our approach in both simulation and on hardware in the real world on a ClearPath TurtleBot using a suite of language-instructed manipulation tasks from the KnowNo Simulation and Mobile Manipulator Datasets \cite{renRobotsThatAsk2023}. We show that LBAP significantly increases the success rate and reduces the amount of help needed as compared to our baselines across different environments and LLMs. Specifically in our real-world testing, we see a decrease in the human help rate of previous methods by over 33\% at a success rate of 70\%, with a similar decrease at most success rates.
    \item We also ablate against various terms in the LBAP Bayesian inference, equations 6-8, showing their benefits individually and superior performance when combined.
    \item We show that using LBAP on the newest, most powerful language models out-of-the-box (GPT-4) outperforms fine-tuning LBAP and prior art \cite{renRobotsThatAsk2023} on the best models available for fine-tuning on this task (GPT-3.5).
\end{enumerate}

\begin{figure*}[ht]
	\begin{center}
        \vspace{0.25cm}
		\includegraphics[width=0.9\linewidth]{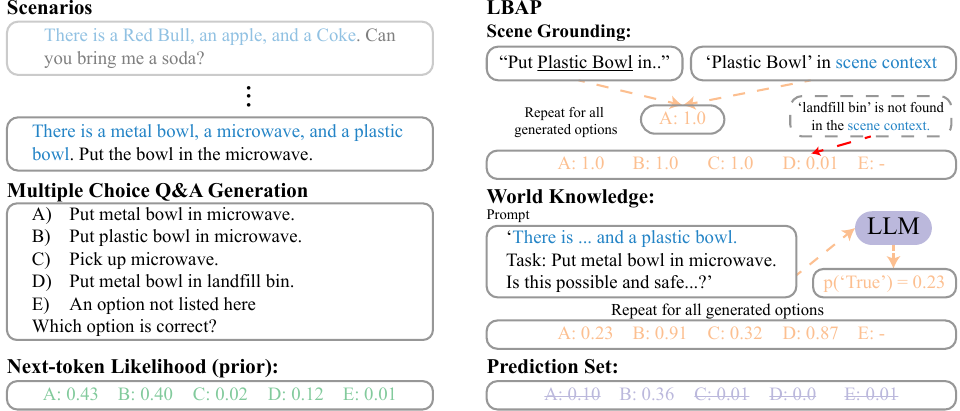}
		\caption{We present LBAP, a novel approach for uncertainty alignment that utilizes Bayesian inference. After generating a set of multiple-choice candidate actions, each prepended with `A', `B', `C', etc., we find the prior using the LLM log probabilities of each option. Using Bayesian inference we then refine these probabilities by determining the grounding of each option in the scene using the scene context and using world knowledge from LLMs to determine the safety and environmental feasibility of each option. After refining the probabilities of the options and renormalizing, we construct a prediction set of options with a probability above a threshold, $t$. Note that we eliminate option A, "Put metal bowl in the microwave," as a valid option as it is dangerous. If the size of the prediction set were greater than one, the agent would ask for help.}
		\label{fig:method}
	\end{center}
\end{figure*}
\section{Related Work}
\subsection{Hallucinations and Uncertainty in LLMs}

As LLMs have risen to the fore, an increasing amount of work has been aimed at addressing their hallucinations, times when the LLM generates content not grounded in reality. Some works aim to quantify the frequency or severity of hallucinations \cite{guanHallusionBenchAdvancedDiagnostic2023} in a given LLM, while many aim to decrease their frequency \cite{dhuliawalaChainofVerificationReducesHallucination2023}, or better calibrate uncertainty \cite{zhouNavigatingGreyArea2023, linTeachingModelsExpress2022, xiaoQuantifyingUncertaintiesNatural2019}. Some recent works attempt to handle hallucinations by using conformal prediction-based methods to provide coverage guarantees\cite{kumarConformalPredictionLarge2023}. Ren et al. \cite{renRobotsThatAsk2023} introduce KnowNo, which uses conformal prediction to create coverage guarantees for robotics tasks, asking users for help when the agent is unsure. In contrast to these works, we improve the confidence measure and mitigate hallucinations through Bayesian inference, minimizing the amount of human help needed. 

\subsection{Action Feasibility and Robotics}

Affordances, or what is possible given environment and object characteristics, is a well-studied topic in computer vision and graphics \cite{do2018affordancenet, hassanPopulating3DScenes2021, affordance2}. Most of these works focus on placing virtual humans or animations into 3D scenes such that the resultant placement generates visually plausible results \cite{hassanPopulating3DScenes2021, mullenPACEDataDrivenVirtual2023, mullenPlacingHumanAnimations2023}. Recent works in robotics use similar techniques to determine what actions are possible and likely for a robot agent given the environmental context, typically for manipulators\cite{yamanobeBriefReviewAffordance2017}. SayCan \cite{ichtersaycan2023} applied these techniques to robot planners by adding an affordance value, obtained from trained value functions and indicating the value of an action in the environment, into the planning scheme. 

Different from SayCan, we compute all values without the need for training or LLM fine-tuning. 
We choose to do this as training capable value functions for every possible action in the real world, as SayCan does, is extremely difficult due to a lack of domain data and vast task diversity. We also believe that SayCan's approach reduces explainability and will hinder the ability to both debug robots in the real-world and explain to their users why they failed. Additionally, we expand upon SayCan in that we incorporate information beyond simply the ability to ``successfully execute each skill," including the safety of executing the action and its conformation to manually set restrictions.

\subsection{LLMs for Robot Planning and Human-Robot Interaction}

Robot motion planning is a well-studied problem in robotics~\cite{lavalle2006planning,canny1988complexity,manocha1992algebraic}. Recently,
Large Language Models (LLMs) have shown an ever-increasing set of capabilities from reasoning and logic~\cite{weiChainofThoughtPromptingElicits2023, kojimaLargeLanguageModels2023, creswellSelectionInferenceExploitingLarge2022} to math and physics \cite{lewkowyczSolvingQuantitativeReasoning2022}. This has extended to robotics, with LLMs improving the state of the art in tasks from high-level planning \cite{ichtersaycan2023, huangLanguageModelsZeroShot2022, wuTidyBotPersonalizedRobot2023, liuLLMEmpoweringLarge2023}, to object goal navigation\cite{gadreCoWsPastureBaselines2022, dorbalaCanEmbodiedAgent2023}. 

Natural language is one of the most popular methods of producing effective human-robot interaction \cite{padmakumarTEAChTaskDrivenEmbodied2022, dialfred,park2019efficient, cvdn} and the advent of LLMs has made dialogue an even more natural medium of communication between a user and their robotic agent. Recent LLM-based robotic works have extended their tasks, incorporating user interaction \cite{huangInnerMonologueEmbodied2022, eqa, renRobotsThatAsk2023} to receive user instructions or to resolve uncertainty.

A key issue in LLM-based robotics is coming up with a set of candidate actions that respond to the user's instruction. Prompting LLMs directly for each valid step in a robot plan has proven to perform poorly on uncertain tasks \cite{murrayCorrectingLengthBias2018}. This becomes a greater issue when we would like to characterize or measure the model's uncertainty. Furthermore, prior art has shown that the LLM itself is a poor scoring function \cite{murrayCorrectingLengthBias2018}, especially when the plan is expressed in natural language. As such, we choose to build off of prior work, specifically KnowNo, and formulate LLM planning as multiple-choice Q\&A (MCQA), a domain LLMs are more familiar with.
\section{Using Bayesian Inference for Uncertainty Alignment}
In this work, we consider the following problem statement: \textit{Given a robotic agent and a user-provided instruction, execute the instruction completely and accurately while resolving any ambiguities as needed through dialogue with the user. Minimize the amount of dialogue.} Specifically, our robot will begin in a set environment, awaiting a user's instruction. The user will provide an instruction, which may be under-specified, at which point our method will need to evaluate whether or not it requires additional information from the user. This problem space was initially explored by \cite{renRobotsThatAsk2023}. 
For example, if the user asks the robot to put a bowl in the microwave, and there is a plastic and metal bowl on the counter, prior art \cite{renRobotsThatAsk2023}, would need to ask the user for clarification as to which bowl they were referring to. However, our goal for this work is to ensure that the robot knows to use the plastic bowl as putting the metal bowl in the microwave would be dangerous. 

Following \cite{renRobotsThatAsk2023}, we choose to transform the problem into an multiple choice question answer (MCQA) task, which allows us to define the problem as next-token prediction -- aligning well with LLM loss functions and training data. To do this, an LLM is provided a few-shot prompt (provided in supplementary materials) that includes possible next steps in a few example scenarios. The LLM in turn generates a set, $\{y_i\}$, of candidate actions that are semantically different for the \textit{current scenario}. Then, the task of choosing among the candidate actions is formatted as a multiple choice question to the LLM, with each option preceded with `A', `B', `C', `D', or `E' as shown in Figure \ref{fig:method}. The probabilities of each possible token (i.e. the multiple choice options `A', `B', `C', ...) then serve as normalized scores, or a prior, that can be utilized by uncertainty quantification methods, like LBAP and KnowNo \cite{renRobotsThatAsk2023}. We utilize these normalized scores as the prior in our Bayesian inference, after which we generate a prediction set of plans from $\{y_i\}$. Please see \cite{renRobotsThatAsk2023} for further rationale on why MCQA should be used for uncertain robot planning.

\textbf{Bayesian Inference.} We choose to use Bayesian inference for uncertainty alignment as it provides a principled framework for updating the confidence estimates from prior knowledge by incorporating multiple sources of additional new information or observed evidence. By applying Bayes' theorem, $P(A|B) \propto P(B|A)P(A)$, the probability of a hypothesis is adjusted based on both initial expectations and new observations, enabling more accurate confidence calibration for a given task. This approach is particularly beneficial for robotic planning tasks where ambiguity arises from environmental perception, natural language user instructions, or LLM hallucination. In this work specifically, Bayesian inference allows us to condition the likelihood of an action on both scene grounding and world knowledge, better aligning the robot's confidence measure with the probability of success and thus mitigating hallucinations and reducing the amount of human intervention needed. Bayesian inference provides a lightweight and explainable mechanism for uncertainty alignment unlike methods requiring extensive training.

\textbf{Method Overview.} We present an overview of our method, LBAP, in Figure \ref{fig:method}. Given a set of candidate actions ${y_i}$ generated by the LLM in the multiple-choice question-answering (MCQA) framework, LBAP aim to estimate the posterior probability
\begin{equation}
    P(y_i|S, W)
\end{equation}
where $P(y_i)$ is the probability that an action is the one the user commanded. $S$ is the scene and $W$ is what we are referring to as `world knowledge,' such as knowing that you cannot put a metal bowl in the microwave.

Using Bayes' theorem (described above), we can express the posterior probability as:
\begin{equation}
    P(y_i|S, W) \propto P(S, W | A)P(y_i)
\end{equation}
where $P(y_i)$ is the prior probability of an action being optimal and $P(S, W | y_i)$ is the likelihood of observing the given scene and world knowledge if the action is optimal.

Assuming independence between the different sources of evidence given $y_i$, we can approximate the likelihood as:
\begin{equation}
    P(S, W | y_i) \approx P(S | y_i)P(W | y_i)
\end{equation}
which makes our final equation:
\begin{equation}
    P(y_i|S, W) \propto P(y_i)P(S | y_i)P(W | y_i).
\end{equation}
To solve this equation, we break it into its different component parts. As mentioned previously LBAP begins with the MCQA paradigm described above, outputting the model's uncalibrated confidence score for each generated action option which we take as the prior $P(y_i)$. 
We then must compute the other two Bayesian terms, $P(S | y_i)$ and $P(W | y_i)$ for each option to calculate the final probability. 

With each option scored, we generate the prediction set of plans, $\mathcal{P}$ as plans with a score above a threshold, $t$
\begin{equation}
    y_i \in \mathcal{P} \text{ if } P(y_i)>t.
\end{equation}
LBAP is certain if this set is a singleton, and queries the user for help if not.
Calculating an optimal $t$ is the main subject of KnowNo \cite{renRobotsThatAsk2023}, which uses conformal prediction to provide statistical guarantees of a certain success rate. Our work differs significantly from KnowNo as we improve $p(y_i)$ with Bayesian inference.
Our experiments are agnostic to the threshold value used, primarily focusing on improving the relationship between success rate and help rate across all thresholds and subsequent success rates. We ablate against the use of $P(S | y_i)$ and $P(W | y_i)$ in our experimentation section.

\textbf{Grounding the Action in the Scene.}
We hypothesize that we can estimate $P(S|y_i)$ by grounding any objects described in $y_i$ in the scene. As in some experiments, and in many of KnowNo's evaluations, we do not have the ability to access the scene directly through perception, we use the textual description of the scene as a fiat. For these scenarios, we use the context of the scene that in theory the robot has already found and provided to the LLM from its perception algorithms. Specifically, to create the prompts for both MCQA and the scoring of the MCQA options, context about the scene was found, we reuse this context. For example, in the simulation environment, there are three blocks and three bowls, one of each is blue, green, and yellow, respectively. The robot's perception algorithms found these objects and conveyed their existence to the agent as part of its initial prompts. 
While perception information is provided, this does not always prevent the LLM from hallucinating new objects into existence, like a `gold bowl.' 
These types of hallucinations become even more prominent when the hallucinated object is in the user command. An example of this can be seen in Figure \ref{fig:coverimg} with the `coffee mug' object. 

To calculate $P(S|y_i)$, the scene context is first parsed into a set of found objects, $\mathcal{O}$. Each proposed action, $y_i$, is then itself parsed for any objects, $o^i$, present. If all of the objects in $y_i$ are in the set of found objects, $a^i_{c} = 1$. Otherwise, $a^i_{c} = \epsilon$, a small probability for rare cases. To continue our example, if $y_i$ included a `blue bowl' and a `yellow block,' then $a^i_{c} = 1$. If $y_i$ included a `gold bowl,' then $a^i_{c} = \epsilon$. 
To formalize this:
\begin{equation}
    P(S|y_i) = 
    \begin{cases}
        1 & \text{if } \forall o^i \in y_i, \, o^i \in \mathcal{O} \\
        \epsilon & \text{otherwise}.
    \end{cases}
\end{equation}

\textbf{Grounding in the Scene with Perception.}
If perception is directly available, and time does not prohibit its use, taking into account any uncertainty in the perception algorithms themselves is valuable. To do this in our experiments, we run an open-vocabulary vision-language model, like CLIP \cite{CLIP} or ViLD \cite{guOpenvocabularyObjectDetection2022}, on the scene observations with each object, $o^i$ (from the option generated by the LLM, $y_i$) as input. 

Specifically, we take the camera images of the scene, $S$, and feed them through ViLD with the list of objects $\{o_i\}$ as input. We then take the probabilities of each object $p(o_i|s)$ as outputted by the model and find the average for a given option, $y_i$. Formally, we define $P(S|y_i)$ as
\begin{equation}
    P(S|y_i) = \prod_{o_i\in y_i}p(o_i|s),
\end{equation}
for $n$ objects in $y_i$. If the object is localized to the same location as an object in the scene context using IoU (usually an indicator of a hallucination), the metric is lowered to $\epsilon$ to prevent duplicate options of the same object.



\textbf{Applying World Knowledge.}
Recent works like \cite{dhuliawalaChainofVerificationReducesHallucination2023} and \cite{weiChainofThoughtPromptingElicits2023} have shown that additional prompts to the LLM requiring it to reason on its prior outputs can isolate harmful hallucinations. We leverage this idea to estimate $P(W|y_i)$, using the world knowledge in LLMs to determine if the generated action conforms to it. This approach also allows for flexibility, where our prompt or sequence of prompts can account for factors beyond just the grounding of objects in the scene, like the safety of completing a given action. Our prompt, given our real-world evaluation environment, begins with a series of few-shot samples, before serving the following text:

\begin{quote}\textit{
\{Header\}
\newline
We: On the counter, there is \{$\mathcal{O}$\}.\newline
We: \{$y_i$\}\newline
We: Is this possible and safe given the provided knowledge of the scene?\newline
You:}
\end{quote}
In the prompt, the items inside the brace notation are replaced with their respective values. The header includes the description of items found in the scene from the MCQA generation and scoring prompts and describes the task for the LLM: determining whether the given action is `possible and safe' in the current scene. The header also tells the LLM to reply in either `True' or `False' and nothing else, as well as 
provides a few examples. Once we receive the response from the LLM, we extract the normalized probability of the `True' token, $p("True"|pr)$ and utilize this as $P(W|y_i)$ for the option $y_i$. Notable is that we believe this approach can be applied across different prompts eliciting different aspects of `world knowledge.' For example, if we have some use case that has business rules such as `Do not leave the living room' we can create a new prompt $pr_{i}$ following the above structure and evaluate that requirement. Creating a set of prompts $PR$ and combining their results gives us:
\begin{equation}
    P(W|y_i) = \prod_{pr_i\in PR}p("True"| pr_i).
\end{equation}
In the version of LBAP we evaluate in our paper, we only employ the provided above prompt. By evaluating these conditions now instead of in the initial prompt to the LLM ensures that they can be evaluated independently, allowing for greater explainability in our overall system and significantly improved performance.

\begin{figure*}[ht]
	\begin{center}
        \vspace{0.25cm}
		\includegraphics[width=\linewidth]{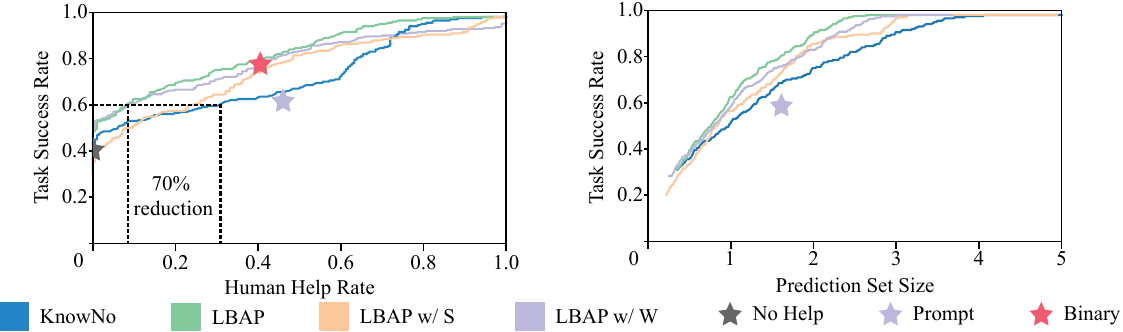}
		\caption{Comparison of task success rate vs human help rate (left) and average prediction set size (right) in our Simulation experiments. Note \textbf{LBAP} improving upon \textbf{KnowNo} for almost all success rates. Including a reduction in the human help rate of over 70\% at a success rate of 60\%. \textbf{No Help} and \textbf{Binary} do not create prediction sets and are not included in the plot.}
		\label{fig:sim-plots}
	\end{center}
\end{figure*}

\textbf{Fine-Tuning.}
We also hypothesize that utilizing the newest, most powerful, LLM available off-the-shelf will generally outperform fine-tuning the older models available for fine-tuning. We tested this hypothesis by fine-tuning \texttt{gpt-3.5-turbo-1106} \cite{gpt3}, the most powerful model available from OpenAI for fine-tuning at the time of our experiments. For the simulation environment, we fine-tuned on each of the three LLM phases of LAP, the MCQA generation, MCQA scoring, and A-Feasibility scoring. 
\section{Experiments and Results}

We evaluate LBAP in a diverse set of language-instructed tasks and environments as described below. Each of these environments is outlined in \cite{renRobotsThatAsk2023}. We chose to use these environments for easy comparison to \cite{renRobotsThatAsk2023} and to continue to move towards a standard set of evaluations for the \textit{uncertainty alignment} task. Each environment has a parameterized scenario distribution with a pre-defined set of possible ambiguities in the human's instruction. We utilize the truth labels that they provide. The full outline of the dataset distribution and every possible ambiguity is included in the supplementary material. We utilize \texttt{GPT-4} \cite{openai2024gpt4} as the LLM in all experiments and ablate against it separately. 


\textbf{Baselines.} Our primary baseline is \textbf{KnowNo} \cite{renRobotsThatAsk2023}, which uses the probability of each option directly from the LLM, and builds a prediction set of options with a probability greater than a threshold value. KnowNo uses conformal prediction to set the threshold value in such a way that a statistical guarantee of a certain success rate is possible. They do nothing to improve the confidence measure itself outside of the task framing as MCQA. No other baseline is known to exist which enables asking the user for help in the MCQA paradigm. All of the additional baselines also use this MCQA framing. Other baselines include:
\begin{itemize}
    \item \textbf{Prompt:} Prompts the LLM to output the prediction set (e.g. ``Prediction set: [B, D]") Exact prompts are in the supplemental materials.
    \item \textbf{Binary:} Prompts the LLM to directly output a binary indicator of uncertainty (e.g. ``Certain/Uncertain: Certain") similar to \cite{huangInnerMonologueEmbodied2022}.
    \item \textbf{No Help:} Always uses the highest score directly from the LLM without creating a prediction set or asking for human intervention.
\end{itemize}

\textbf{Ablations.} To test the necessity of each term in our system, we ablate against the use of the scene grounding, $S$, and world knowledge, $W$, in LBAP. Specifically, the ablations include:
\begin{itemize}
    \item \textbf{LBAP w/ $S$}: We evaluate $P(y_i|S)$ instead of the full posterior probability. This essentially removes the $P(W | y_i)$ term.
    \item \textbf{LBAP w/ $W$}: We evaluate $P(y_i|W)$ instead of the full posterior probability. This essentially removes the $P(S | y_i)$ term.
\end{itemize}

The \textbf{LBAP w/ $S$} ablation is similar to SayCan \cite{ichtersaycan2023}, but using active perception instead of pre-trained value functions for each possible action. The results for these ablations are described in the main experimentation sections. We also ablate against the specific LLM used, changing the LLM for only the MCQA, and fine-tuning the LLM and describe these experiments in a different subsection.

\textbf{Evaluation Metrics.} We primarily use human-help rate and prediction set size versus success rate to compare LBAP to baselines. Intuitively, when given a set success rate, these metrics provide information about the help rate and average prediction set size for that success rate, with lower help rates and lower average prediction set sizes better. For a given data sample, the human-help rate is 1 if the prediction set size is greater than 1.
\textbf{Prompt} and \textbf{Binary} produce a single success rate and thus a singular help rate and average prediction set size. With KnowNo and LBAP, we have a tunable threshold value that is used to confine the prediction set. For these methods, we apply threshold values, $t$, ranging from 0.0000001 to 0.7 for full coverage. The success rate and human help rate or average prediction set size for each threshold value are then plotted, and the area under the resultant curve (AuC) becomes a valuable metric to quantify an improved relationship between task success rate and human help rate.

\begin{figure*}[ht]
	\begin{center}
        \vspace{0.25cm}
		\includegraphics[width=\linewidth]{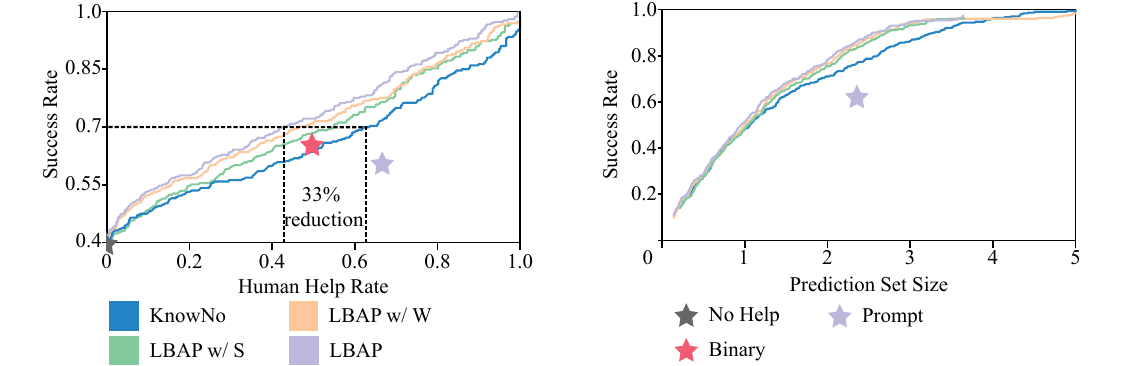}
		\caption{Comparison of task success rate vs human help rate (left) and prediction set size (right) in our real-world experiments on a ClearPath TurtleBot. Note the improvement over \textbf{KnowNo} across all success rates. \textbf{LBAP} improves upon \textbf{KnowNo} and its ablations with just $W$ or $S$ in the Bayesian inference.}
		\label{fig:rw-plots}
	\end{center}
\end{figure*}

\subsection{Simulation: Tabletop Rearrangement}
In this task, a robot arm is asked to rearrange objects on a table in the PyBullet simulator. Each scenario is initialized with three bowls and three blocks, with one each of blue, green, and yellow, respectively. The task requires the robot to move a certain number of blocks or bowls toward a specific location relative to a different object. For example, ``Move the Green Block to the left of the Blue Bowl." As introduced KnowNow, three different ambiguities are injected into the user instruction: (1) \textit{attribute}, where `block' or `bowl' is replaced with a potentially ambiguous term like `thing' or `cube,' (2) \textit{numeric}, where the number of blocks to move is underspecified by using a term like `some' in place of `two,' and (3) \textit{spatial}, where the exact location like `left' or ` front' is replaced with an ambiguous term like `near.' The scenario distribution is uniform over all possible ambiguities and a testing set is generated for use. All LBAP methods ground the actions using our perception based scoring.

We show in Figure \ref{fig:sim-plots} that \textbf{LBAP} outperforms \textbf{KnowNo}, as well as all other baselines and ablations, with a higher success-to-help ratio across threshold values. Quantitatively, \textbf{LBAP} reduces the amount of help needed by over 70\%, from 30.1\% with \textbf{KnowNo} to 8.9\% at a success rate of 60\%. 
Similarly, prediction sets created by \textbf{LBAP} are smaller than those from \textbf{KnowNo}. 
On the Task~Success~Rate~$\times$~Human~Help~Rate curve, we find that \textbf{LBAP} has an Auc of $0.780$ compared to \textbf{KnowNo} with an AuC of $0.699$.

Of \textbf{Prompt} and \textbf{Binary}, both produce a singular success rate and help rate. Binary performed the best for this environment, getting close to the performance of \textbf{LBAP} at its success level of 77\%. \textbf{Prompt} performs relatively poorly, with a success rate below 60\%, illustrating the challenges present when attempting to retrieve the prediction sets directly from the LLM. 

LBAP with scene grounding, \textbf{LBAP w/ $S$} overperformed LBAP without scene knowledge \textbf{LBAP w/ $W$}. We believe this is caused by MCQA options, $y^i$, becoming more difficult to separate relative to \textbf{LBAP w/ $S$}, which uses more aggressive probabilities for filtering out hallucinations in from the action options. However, the combined method LBAP with the full Bayesian inference shows the best results, indicating that combining world knowledge \textit{and} scene information are important for optimal performance. 

\begin{figure}[t]
	\begin{center}
		\includegraphics[width=\columnwidth]{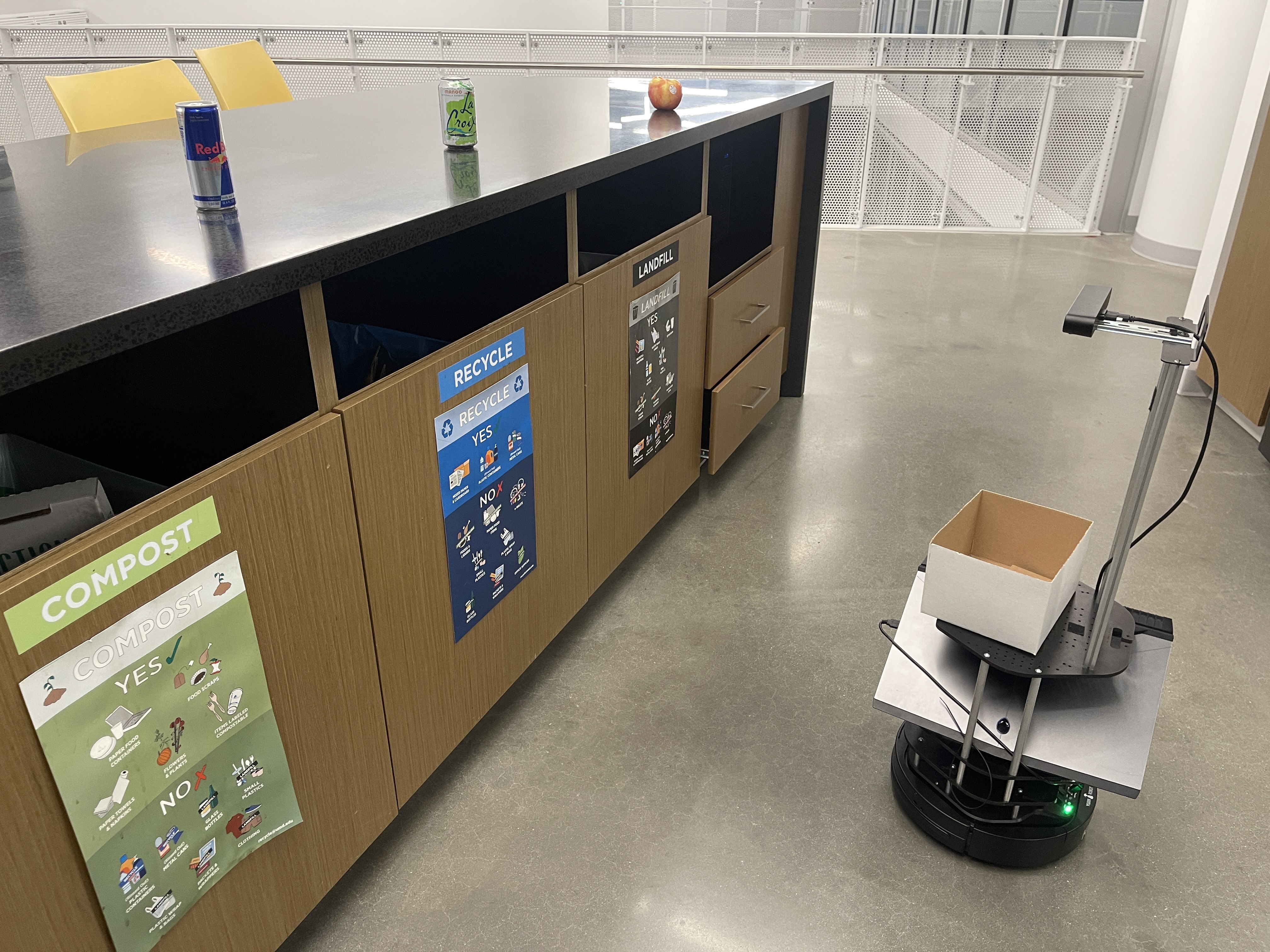}
		\caption{Example of our real-world experimentation setup. The ClearPath TurtleBot must navigate towards the correct object, at which point a human moves the object into the basket before the robot continues towards a goal location. In this specific example, the perception has grounded a `Red Bull,' a `soda,' and an `apple.' We notice frequent instances of a hallucination caused by the user, for example, the user asking for a `fruit' causing the agent to hallucinate a `fruit' object which is not properly grounded in the scene, instead of the agent selecting the `apple.'}
		\label{fig:real-world}
	\end{center}
    \vspace{-0.4cm}
\end{figure}

\subsection{Real World: Mobile Manipulator in a Kitchen}
For this environment, we continue to use the task specifications laid out in KnowNo \cite{renRobotsThatAsk2023}. Shown in Figure \ref{fig:real-world}, each scenario involves a mobile robot in front of a countertop in an office kitchen, next to a set of recycling/compost/landfill bins. The tasks include picking up objects from the countertop and possibly putting it into a bin, or somewhere else on the countertop. New ambiguities are added relative to the simulator environment, including some involving unsafe actions e.g. task: ``place the bowl in the microwave”  context: In the scene there is a plastic bowl and a metal bowl.
The full outline of every possible ambiguity is included in the supplementary material. In our lab, no viable mobile manipulator exists, so we run our testing on a ClearPath TurtleBot. During the testing, we manually pick up the object selected by the agent and put it in a basket on the robot, simulating the use of the manipulator. We mark a task a success if the robot indicates the correct actions, moves to within 1 meter of the target object or destination, and orients itself towards the target. 

Similar to the simulation environment, we find that \textbf{LBAP} outperforms \textbf{KnowNo} and our other baselines. We show these results in Figure \ref{fig:rw-plots}. \textbf{LBAP} indicates with a human help rate of only 42\% relative to \textbf{KnowNo}'s 63\% at a 70\% planning success rate. 
These findings are corroborated with \textbf{LBAP} having a lower average prediction set size than \textbf{KnowNo}, and an improved AuC of $0.72$ relative to \textbf{KnowNo}'s $0.67$. Qualitatively, we notice that \textbf{LBAP} improves performance on scenarios with a safety component, likely due to the incorporation of world knowledge $W$ evaluating each option partially on its safety. This is backed up by the \textbf{LBAP w/ $W$} ablation performing better than \textbf{LBAP w/ $S$}.
We can identify a similar trend for the \textbf{Binary} and \textbf{Prompt} baselines as in the simulation environment, with \textbf{Binary} outperforming \textbf{Prompt}, but with neither getting particularly close to \textbf{LBAP}. \textbf{Prompt} performs worse relative to the other methods in the real-world environment, exhibiting how the added complexity and ambiguities of the environment further hampering the \textbf{Prompt} baseline's ability to produce an accurate prediction set directly. 

We test \textbf{LBAP w/ $W$} and \textbf{LBAP w/ $S$} finding that each improves performance over \textbf{KnowNo} individually. However, combining them in \textbf{LBAP} allows for the greatest effect. We believe this validates our Bayesian inference with each term contributing different things to the final probability. $W$ measures not just possibility, but also safety. Meanwhile, $S$ does a very good job at removing any hallucinated objects or destinations from the list of available options entirely.


\begin{figure}[t]
	\begin{center}
        \vspace{0.25cm}
		\includegraphics[width=\columnwidth]{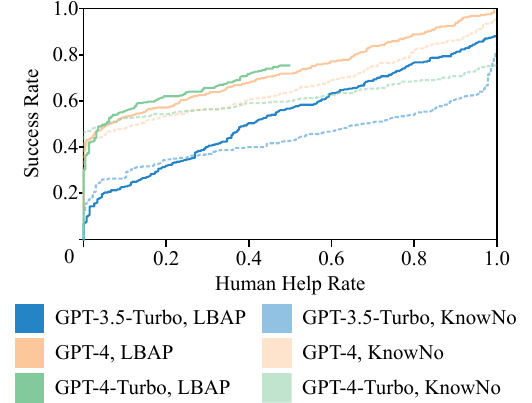}
		\caption{Comparison of \textbf{LBAP} and \textbf{KnowNo} across different LLMs. Note that \textbf{LBAP} improves upon \textbf{KnowNo} with every model, and that performance generally increases with each model. We noticed qualitatively that \texttt{GPT-4-Turbo} exhibited poor performance in the MCQA generation phase relative to \texttt{GPT-4}, frequently asking for clarifications instead of producing a set of candidate options, capping the possible success rate.}
		\label{fig:model-ablate}
	\end{center}
    \vspace{-0.4cm}
\end{figure}

\subsection{LLM Ablations}
To further test \textbf{LBAP}, we ablated our model choice, \texttt{GPT-4} for both our method, \textbf{LBAP}, and \textbf{KnowNo} in our real-world paradigm. Figure \ref{fig:model-ablate} shows these results. We show a large improvement over \textbf{KnowNo} across all models. The performance gap between \textbf{LBAP} and \textbf{KnowNo} is wider for \texttt{GPT-3.5-Turbo} and \texttt{GPT-4-Turbo} than it is for \texttt{GPT-4}. We hit maximum performance on \texttt{GPT-4-Turbo} at half the human help rate of \textbf{KnowNo}. Additionally, we show that the performance of \textbf{LBAP} improves with the model. This encouraging behavior implies that when models get released in the future, \textbf{LBAP} will continue to apply and increase their performance for these tasks.


As an additional ablation, we also tested \textbf{LBAP} using the ablated model for the MCQA generation and scoring phases, but only \texttt{GPT-3.5-Turbo} for evaluating the world knowledge. Performance was largely similar with help rate increasing by under 4\% on average. We envision this as a cost-saving measure for users of \textbf{LBAP} who want to minimize API or compute costs. 
These plots can be seen in our supplementary materials.

\subsection{Why Not Just Fine-Tune?}
In simulation, where we can generate a large set of possible tasks, we attempted to fine-tune \texttt{gpt-3.5-turbo-1106}, the best model available at the time, on each part of \textbf{LBAP}. 
To test generalization, we separate out certain ambiguities of each type, as well as one entire type, for use exclusively in the testing set, which has a distribution identical to our dataset utilized above. 

We find that overall help rates increase for the same model by on average 14\%. This difference is largely due to impact on performance for ambiguities that were outside of the training data, showing overfitting as seen by others including \cite{renRobotsThatAsk2023}. Similar issues occurred when fine-tuning on the option scoring component of \textbf{LBAP}, decreasing help rates by less than 5\% on average, but requiring help more frequently on the withheld classes. This method underperforms \textbf{LBAP} on GPT-4 across most success rates. However, prediction set sizes were decreased more than \textbf{LBAP} on average, showing the model improving its rejection of poor options.

In the real-world environment, with even more limited data, fine-tuning the MCQA proposal produced an overfitted model that could not generalize to the testing set of tasks.
When fine-tuning on the scoring task, we saw a similar phenomenon to in the simulation paradigm with improved performance on samples inside the training distribution, but poorer performance on those outside. 
When finetuning on scoring, help rates decreased on average but with significantly worse performance in withheld categories.
The fine-tuned models still underperform \textbf{LBAP} on GPT-4. 
We can draw two separate conclusions from these experiments: (1) Generating a large and diverse enough dataset of scenarios in a robotic paradigm is exceptionally difficult, and (2) The newest, most powerful, models available outperform a fine-tuned older model for this task.

\section{Conclusions, Limitations, and Future Work}
Robots must complete user-provided instructions accurately, with minimal instruction from the user. In this work we introduce LBAP, a method of using Bayesian inference to significantly reduce the impact of LLM hallucinations and minimize the amount of human feedback needed when completing tasks. This includes a reduction in human help over prior art KnowNo \cite{renRobotsThatAsk2023} of 33\% at a success rate of 70\% in our real-world experiments. LBAP presents a strong improvement over the confidence score used in KnowNo in all cases while retaining the ability to use conformal prediction to set up statistical performance guarantees.

\textbf{Limitations and Future Work.} 
The primary limitation is the limited diversity of the dataset we tested on, mainly that it only includes pick-and-place tasks. Future work on more diverse datasets would be valuable. This would especially help to validate the incorporation of world knowledge to tasks that involve rules or restrictions that are not otherwise obvious but could be incorporated into the prompts.




{\small
\bibliographystyle{IEEEtran}
\bibliography{refs}

\begin{thebibliography}{10}
\providecommand{\url}[1]{#1}
\csname url@samestyle\endcsname
\providecommand{\newblock}{\relax}
\providecommand{\bibinfo}[2]{#2}
\providecommand{\BIBentrySTDinterwordspacing}{\spaceskip=0pt\relax}
\providecommand{\BIBentryALTinterwordstretchfactor}{4}
\providecommand{\BIBentryALTinterwordspacing}{\spaceskip=\fontdimen2\font plus
\BIBentryALTinterwordstretchfactor\fontdimen3\font minus \fontdimen4\font\relax}
\providecommand{\BIBforeignlanguage}[2]{{%
\expandafter\ifx\csname l@#1\endcsname\relax
\typeout{** WARNING: IEEEtran.bst: No hyphenation pattern has been}%
\typeout{** loaded for the language `#1'. Using the pattern for}%
\typeout{** the default language instead.}%
\else
\language=\csname l@#1\endcsname
\fi
#2}}
\providecommand{\BIBdecl}{\relax}
\BIBdecl

\bibitem{ichtersaycan2023}
B.~Ichter, A.~Brohan \emph{et~al.}, ``Do {{As I Can}}, {{Not As I Say}}: {{Grounding Language}} in {{Robotic Affordances}},'' in \emph{Proceedings of {{The}} 6th {{Conference}} on {{Robot Learning}}}.\hskip 1em plus 0.5em minus 0.4em\relax {PMLR}, Mar. 2023, pp. 287--318.

\bibitem{dorbalaCanEmbodiedAgent2023}
V.~S. Dorbala, J.~F.~M. Jr, and D.~Manocha, ``Can an {{Embodied Agent Find Your}} ``{{Cat-shaped Mug}}''? {{LLM-Based Zero-Shot Object Navigation}},'' \emph{IEEE Robotics and Automation Letters}, pp. 1--8, 2023.

\bibitem{renRobotsThatAsk2023}
A.~Z. Ren, A.~Dixit \emph{et~al.}, ``Robots {{That Ask For Help}}: {{Uncertainty Alignment}} for {{Large Language Model Planners}},'' Sep. 2023.

\bibitem{guanHallusionBenchAdvancedDiagnostic2023}
T.~Guan, F.~Liu \emph{et~al.}, ``{{HallusionBench}}: {{An Advanced Diagnostic Suite}} for {{Entangled Language Hallucination}} \& {{Visual Illusion}} in {{Large Vision-Language Models}},'' Nov. 2023.

\bibitem{dhuliawalaChainofVerificationReducesHallucination2023}
S.~Dhuliawala, M.~Komeili \emph{et~al.}, ``Chain-of-{{Verification Reduces Hallucination}} in {{Large Language Models}},'' Sep. 2023.

\bibitem{huangInnerMonologueEmbodied2022}
W.~Huang, F.~Xia \emph{et~al.}, ``Inner {{Monologue}}: {{Embodied Reasoning}} through {{Planning}} with {{Language Models}},'' Jul. 2022.

\bibitem{dialfred}
X.~Gao, Q.~Gao \emph{et~al.}, ``Dialfred: Dialogue-enabled agents for embodied instruction following,'' \emph{IEEE Robotics and Automation Letters}, vol.~7, no.~4, pp. 10\,049--10\,056, 2022.

\bibitem{cvdn}
J.~Thomason, M.~Murray \emph{et~al.}, ``Vision-and-dialog navigation,'' in \emph{Conference on Robot Learning}.\hskip 1em plus 0.5em minus 0.4em\relax PMLR, 2020, pp. 394--406.

\bibitem{zhouNavigatingGreyArea2023}
K.~Zhou, D.~Jurafsky, and T.~Hashimoto, ``Navigating the {{Grey Area}}: {{How Expressions}} of {{Uncertainty}} and {{Overconfidence Affect Language Models}},'' in \emph{EMNLP}, 2023, pp. 5506--5524.

\bibitem{linTeachingModelsExpress2022}
S.~Lin, J.~Hilton, and O.~Evans, ``Teaching {{Models}} to {{Express Their Uncertainty}} in {{Words}},'' Jun. 2022.

\bibitem{xiaoQuantifyingUncertaintiesNatural2019}
Y.~Xiao and W.~Y. Wang, ``Quantifying {{Uncertainties}} in {{Natural Language Processing Tasks}},'' \emph{Proceedings of the AAAI Conference on Artificial Intelligence}, vol.~33, no.~01, pp. 7322--7329, Jul. 2019.

\bibitem{kumarConformalPredictionLarge2023}
B.~Kumar, C.~Lu \emph{et~al.}, ``Conformal {{Prediction}} with {{Large Language Models}} for {{Multi-Choice Question Answering}},'' Jul. 2023.

\bibitem{do2018affordancenet}
T.-T. Do, A.~Nguyen, and I.~Reid, ``Affordancenet: An end-to-end deep learning approach for object affordance detection,'' in \emph{ICRA}, 2018.

\bibitem{hassanPopulating3DScenes2021}
M.~Hassan, P.~Ghosh \emph{et~al.}, ``Populating {{3D Scenes}} by {{Learning Human-Scene Interaction}},'' in \emph{2021 {{IEEE}}/{{CVF Conference}} on {{Computer Vision}} and {{Pattern Recognition}} ({{CVPR}})}.\hskip 1em plus 0.5em minus 0.4em\relax {IEEE}, 2021.

\bibitem{affordance2}
X.~Li, S.~Liu \emph{et~al.}, ``Putting humans in a scene: Learning affordance in 3d indoor environments,'' in \emph{Proceedings of the IEEE/CVF Conference on Computer Vision and Pattern Recognition (CVPR)}, June 2019.

\bibitem{mullenPACEDataDrivenVirtual2023}
J.~F. Mullen and D.~Manocha, ``{{PACE}}: {{Data-Driven Virtual Agent Interaction}} in {{Dense}} and {{Cluttered Environments}},'' \emph{IEEE Transactions on Visualization and Computer Graphics}, May 2023.

\bibitem{mullenPlacingHumanAnimations2023}
J.~F. Mullen, D.~Kothandaraman \emph{et~al.}, ``Placing {{Human Animations Into 3D Scenes}} by {{Learning Interaction-}} and {{Geometry-Driven Keyframes}},'' in \emph{Proceedings of the {{IEEE}}/{{CVF Winter Conference}} on {{Applications}} of {{Computer Vision}}}, 2023, pp. 300--310.

\bibitem{yamanobeBriefReviewAffordance2017}
N.~Yamanobe, W.~Wan \emph{et~al.}, ``A brief review of affordance in robotic manipulation research,'' \emph{Advanced Robotics}, vol.~31, no. 19-20, pp. 1086--1101, Oct. 2017.

\bibitem{lavalle2006planning}
S.~LaValle, ``Planning algorithms,'' \emph{Cambridge University Press google schola}, vol.~2, pp. 3671--3678, 2006.

\bibitem{canny1988complexity}
J.~Canny, \emph{The complexity of robot motion planning}.\hskip 1em plus 0.5em minus 0.4em\relax MIT press, 1988.

\bibitem{manocha1992algebraic}
D.~Manocha, \emph{Algebraic and numeric techniques in modeling and robotics}.\hskip 1em plus 0.5em minus 0.4em\relax University of California, Berkeley, 1992.

\bibitem{weiChainofThoughtPromptingElicits2023}
J.~Wei, X.~Wang \emph{et~al.}, ``Chain-of-{{Thought Prompting Elicits Reasoning}} in {{Large Language Models}},'' Jan. 2023.

\bibitem{kojimaLargeLanguageModels2023}
T.~Kojima, S.~S. Gu \emph{et~al.}, ``Large {{Language Models}} are {{Zero-Shot Reasoners}},'' Jan. 2023.

\bibitem{creswellSelectionInferenceExploitingLarge2022}
A.~Creswell, M.~Shanahan, and I.~Higgins, ``Selection-{{Inference}}: {{Exploiting Large Language Models}} for {{Interpretable Logical Reasoning}},'' in \emph{The {{Eleventh International Conference}} on {{Learning Representations}}}, Sep. 2022.

\bibitem{lewkowyczSolvingQuantitativeReasoning2022}
A.~Lewkowycz, A.~Andreassen \emph{et~al.}, ``Solving {{Quantitative Reasoning Problems}} with {{Language Models}},'' Jun. 2022.

\bibitem{huangLanguageModelsZeroShot2022}
W.~Huang, P.~Abbeel \emph{et~al.}, ``Language {{Models}} as {{Zero-Shot Planners}}: {{Extracting Actionable Knowledge}} for {{Embodied Agents}},'' in \emph{Proceedings of the 39th {{International Conference}} on {{Machine Learning}}}.\hskip 1em plus 0.5em minus 0.4em\relax {PMLR}, Jun. 2022, pp. 9118--9147.

\bibitem{wuTidyBotPersonalizedRobot2023}
J.~Wu, R.~Antonova \emph{et~al.}, ``{{TidyBot}}: Personalized robot assistance with large language models,'' \emph{Autonomous Robots}, vol.~47, no.~8, pp. 1087--1102, Nov. 2023.

\bibitem{liuLLMEmpoweringLarge2023}
B.~Liu, Y.~Jiang \emph{et~al.}, ``{{LLM}}+{{P}}: {{Empowering Large Language Models}} with {{Optimal Planning Proficiency}},'' Sep. 2023.

\bibitem{gadreCoWsPastureBaselines2022}
S.~Y. Gadre, M.~Wortsman \emph{et~al.}, ``{{CoWs}} on {{Pasture}}: {{Baselines}} and {{Benchmarks}} for {{Language-Driven Zero-Shot Object Navigation}},'' Dec. 2022.

\bibitem{padmakumarTEAChTaskDrivenEmbodied2022}
A.~Padmakumar, J.~Thomason \emph{et~al.}, ``{{TEACh}}: {{Task-Driven Embodied Agents That Chat}},'' \emph{Proceedings of the AAAI Conference on Artificial Intelligence}, vol.~36, no.~2, pp. 2017--2025, Jun. 2022.

\bibitem{park2019efficient}
J.~S. Park, B.~Jia \emph{et~al.}, ``Efficient generation of motion plans from attribute-based natural language instructions using dynamic constraint mapping,'' in \emph{ICRA}.\hskip 1em plus 0.5em minus 0.4em\relax IEEE, 2019.

\bibitem{eqa}
V.~S. Dorbala, P.~Goyal \emph{et~al.}, ``S-eqa: Tackling subjective queries in embodied question answering,'' \url{https://gamma.umd.edu/researchdirections/embodied/seqa/}, [Accessed 26-02-2024].

\bibitem{murrayCorrectingLengthBias2018}
K.~Murray and D.~Chiang, ``Correcting {{Length Bias}} in {{Neural Machine Translation}},'' in \emph{Proceedings of the {{Third Conference}} on {{Machine Translation}}: {{Research Papers}}}, Oct. 2018, pp. 212--223.

\bibitem{CLIP}
A.~Radford, J.~W. Kim \emph{et~al.}, ``Learning transferable visual models from natural language supervision,'' in \emph{International conference on machine learning}.\hskip 1em plus 0.5em minus 0.4em\relax PMLR, 2021, pp. 8748--8763.

\bibitem{guOpenvocabularyObjectDetection2022}
X.~Gu, T.-Y. Lin \emph{et~al.}, ``Open-vocabulary {{Object Detection}} via {{Vision}} and {{Language Knowledge Distillation}},'' May 2022.

\bibitem{gpt3}
T.~Brown, B.~Mann \emph{et~al.}, ``Language models are few-shot learners,'' \emph{Advances in neural information processing systems}, 2020.

\bibitem{openai2024gpt4}
OpenAI, J.~Achiam \emph{et~al.}, ``Gpt-4 technical report,'' 2024.

\end{thebibliography}
}


\newpage
\appendices

\section{Additional Experiment Details}
\textbf{Full Prompt for Prompt-Based A-Feasibility}
\begin{quote}\textit{
\{Header\}\newline
...
\newline\newline
We: On the counter, there is a metal bowl, a plastic bowl, and a microwave.\newline
We: pick up the metal bowl and put it in the microwave\newline
We: Is this possible and safe given the provided knowledge of the scene?\newline
You: False
\newline\newline
We: On the counter, there is an orange, a bag of rice chips, and an apple.\newline
We: pick up the orange\newline
We: Is this possible and safe given the provided knowledge of the scene?\newline
You: True
\newline\newline
We: On the counter, there is \{$\mathcal{O}$\}.\newline
We: \{$y^i$\}\newline
We: Is this possible and safe given the provided knowledge of the scene?\newline
You:}
\newline
\end{quote}

\textbf{MCQA Prompt.}
ADD HERE

\textbf{Simulation setting.} This information on the simulated task is directly from \cite{renRobotsThatAsk2023}, the source of the task.

\begin{itemize}
    \item  Environment: there are always three blocks and bowls of color red, yellow, and green with random locations on the table.
    \item  Goal: we use the following template: {put, place, move} {a, one, a single of, two, a pair of, three, all, red, yellow, green} {block(s), bowl(s)} {on, to the left of, to the right of, to the front of, at the back of} the {red, green, yellow} {block(s), bowl(s)}. The scenario distribution is uniform over these possibilities.
    \item Instruction: for the language instructions, we modify the goal (from the template) according to the ambiguity type. The scenario distribution is uniform over the listed ambiguous cases in each ambiguity type.
    \begin{itemize}
        \item Attribute ambiguities: refer to the block as one of “cube”, “cuboid”, “box”, “square object”, to the bowl as one of “container”, “round object”, “receptacle”, or to either block or bowl as one of “object”, “item”, “thing” (“move the blue object in yellow bowl”); refer to “blue” as one of
        “cyan”, “navy”, to “green” as one of “greenish”, “grass-colored”, and to “yellow” as “orange” or “gold”. This setting is the least ambiguous one among the three ambiguity types.
        \item Numeric ambiguities: refer to either two or three numerically with one of “a few”, “a couple of”, “some”, “a handful of” (“put some blocks in the green bowl”).
        \item Spatial ambiguities: refer to any of the four possible directions with “near”, “close to”,“beside”, “next to”, refer to either left to right with “lateral to”, and refer to either front or behind with “along the line of sight”. This setting is the most ambiguous one among the three ambiguity types.
    \end{itemize}
\end{itemize}

Outsize of \cite{renRobotsThatAsk2023}'s task description, we exchanged communications with them that provided further clarification for setting up this task. Our additional findings which should help with setting up this task are below.
\begin{itemize}
    \item For goals with numeric ambiguities, the `block' is always the object to move, and all the `blocks' are the same color to avoid multiple ambiguities.
    \item The authors of \cite{renRobotsThatAsk2023} used different few shot prompts for each ambiguity, so that the examples provided matched the ambiguity accordingly.
\end{itemize}

We chose not to conform to the second point, instead using one few shot prompt for all ambiguities. We believe that this is more faithful to the task as the agent should have no knowledge of which ambiguity is being provided to it. We did test multiple prompts with varying examples and chose the one with the best results, although results did not change much when using reasonable examples of all ambiguity types. Our code, which will be release after acceptance, will show how we produced the full dataset listing from \cite{renRobotsThatAsk2023} and should help resolve any additional questions.

\textbf{Hardware Mobile Manipulator setting.} This information on the mobile manipulator task is directly from \cite{renRobotsThatAsk2023}, the source of the task.

\begin{itemize}
    \item Environment: the full list of possible objects include: bottled water, bottled tea, orange soda, RedBull, Coke, Pepsi, Sprite, rice chips, jalapeno chips, kettle chips, multigrain chips, apple, orange, energy bar, clean sponge, dirty sponge, metal bowl, plastic bowl. Depending on the ambiguity listed below, there is three objects placed on the top of the counter (including randomly sampled distractors from the list). There is also a set of landfill, compost, and recycling bins, a microwave, and a portable stove.
    \item Instruction: for convenience, we introduce the possible instructions first in different ambiguous scenarios; they each correspond to possible goals. Please refer to \url{https://robot-help.github.io/prompts/mobile_tasks.txt} for the full list. The possible instructions are a uniform distribution over different types: (1) single-label, e.g., ‘Bring me a Coke’ (unambiguous); (2) creative-single-label, e.g., ‘I want a healthy fruit to munch on.’ which means the apple (unambiguous); (3) multi-label, e.g., ‘Bring me a cola.’ and either Coke or Pepsi is acceptable; (4) creative-multi-label, e.g., ‘Bring me something with a kick.’ and either RedBull or jalapeno chips are acceptable; (5) spatially-ambiguous, e.g., ‘Put the Coke in the drawer’ or ‘Put the Coke near the fruit’ which under-specifies the drawer or fruit; (6) unsafe, e.g., ‘Can you dispose of the bottle drink? It should have expired.’ or ‘Place the bowl on the stove, please.’; (7) Winograd, e.g., ’There is a sponge and a bag of rice chips...I don’t want to use it for cleaning any more. Can you please dispose of it?” We use the GPT-4 model for generating the creative tasks.
    \item Goal: the corresponding goal for the ambiguous instructions above. For example, the instruction is “Put the Coke in the drawer”, and the goal is uniform over the two possibilities: put the Coke in the
    top drawer, and put the Coke in the bottom drawer.
\end{itemize}

We faithfully implement \cite{renRobotsThatAsk2023}'s task description exactly as described, using their code provided here: \url{https://github.com/google-research/google-research/tree/master/language_model_uncertainty}. They specifically provide two files with their 300 tasks and corresponding prompts in the code. While this task is challenging and diverse, we believe that more can be done to expand the task to a much larger set of commands and situations. This is very limited in scope with very few object classes, and most commands boiling down to picking up and moving an object. Future work must be conducted to expand this task for more general future robots.

\section{Additional LLM Ablations}
\begin{figure}[t]
	\begin{center}
		\includegraphics[width=.9\columnwidth]{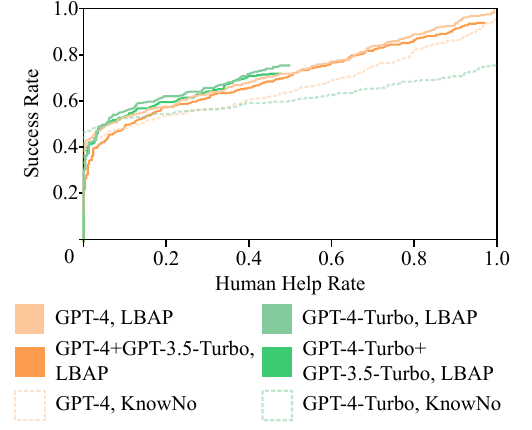}
		\caption{Comparison of \textbf{LAP+$a_{pr}\times a_c$} and \textbf{KnowNo} accross different LLMs. Note that \textbf{LAP} improves upon \textbf{KnowNo} with every model, and that performance overall generally increases with each model. \texttt{GPT-4-Turbo} exhibited poor performance in the MCQA generation phase relative to \texttt{GPT-4}, frequently asking for clarifications instead of producing a set of candidate options, capping the possible success rate.}
		\label{fig:model-ablate2}
	\end{center}
    \vspace{-0.5cm}
\end{figure}

As described in our experimentation section, we tested \textbf{LAP} with one model (\texttt{GPT-4}, \texttt{GPT-4-Turbo}, and \texttt{GPT-3.5-Turbo}) for the MCQA generation and scoring, but \texttt{GPT-3.5-Turbo} for finding our prompt-based A-Feasibility score, $a_{pr}$. This is a possible cost saving measure for users of \textbf{LAP} who need to minimize their API or compute costs. This plot is found in Figure \ref{fig:model-ablate2}. We find that performance slightly decreases when using \texttt{GPT-3.5-Turbo} for our prompt-based A-Feasibility score. We suggest that users run some testing on their own use case to determine the right trade off.

\end{document}